\documentclass[conference]{IEEEtran}
\IEEEoverridecommandlockouts
\usepackage{cite}
\usepackage{amsmath,amssymb,amsfonts}
\usepackage{algorithmic}
\usepackage{graphicx}
\usepackage{textcomp}
\usepackage{xcolor}
\usepackage[shortlabels]{enumitem}
\usepackage{url}
\usepackage{balance}
\usepackage{hyperref}

\def\BibTeX{{\rm B\kern-.05em{\sc i\kern-.025em b}\kern-.08em
    T\kern-.1667em\lower.7ex\hbox{E}\kern-.125emX}}
\begin{document}

\title{Enhancing Crime Scene Investigations through Virtual Reality and Deep Learning Techniques
}

\author{
    \IEEEauthorblockN{Antonino Zappalà\IEEEauthorrefmark{1}, Luca Guarnera\IEEEauthorrefmark{1},
    Vincenzo Rinaldi\IEEEauthorrefmark{2}, Salvatore Livatino\IEEEauthorrefmark{3} and
    Sebastiano Battiato\IEEEauthorrefmark{1}}
    \IEEEauthorblockA{\IEEEauthorrefmark{1}Department of Mathematics and Computer Science, University of Catania, Catania, Italy\\
    \{antonino.zappala, luca.guarnera, sebastiano.battiato\}@unict.it}
    \IEEEauthorblockA{\IEEEauthorrefmark{2}Leverhulme Research Centre for Forensic Science, School of Science and Engineering, University of Dundee, Dundee, Scotland\\
    \{vrinaldi001\}@dundee.ac.uk}
    \IEEEauthorblockA{\IEEEauthorrefmark{3}School of Physics, Engineering and Computer Science, University of Hertfordshire, Hatfield, UK\\
    \{s.livatino\}@herts.ac.uk}
}

\maketitle

\begin{abstract}

The analysis of a crime scene is a pivotal activity in forensic investigations. Crime Scene Investigators and forensic science practitioners rely on best practices, standard operating procedures, and critical thinking, to produce rigorous scientific reports to document the scenes of interest and meet the quality standards expected in the courts.
However, crime scene examination is a complex and multifaceted task often performed in environments susceptible to deterioration, contamination, and alteration, despite the use of contact-free and non-destructive methods of analysis. In this context, the documentation of the sites, and the identification and isolation of traces of evidential value remain challenging endeavours.
    In this paper, we propose a photogrammetric reconstruction of the crime scene for inspection in virtual reality (VR) and focus on fully automatic object recognition with deep learning (DL) algorithms through a client-server architecture. A pre-trained Faster-RCNN model was chosen as the best method that can best categorize relevant objects at the scene, selected by experts in the VR environment. These operations can considerably improve and accelerate crime scene analysis and help the forensic expert in extracting measurements and analysing in detail the objects under analysis.
    Experimental results on a simulated crime scene have shown that the proposed method can be effective in finding and recognizing objects with potential evidentiary value, enabling timely analyses of crime scenes, particularly those with health and safety risks (e.g. fires, explosions, chemicals, etc.), while minimizing subjective bias and contamination of the scene.

\end{abstract}


\section{Introduction}
\label{sec:intro}

\begin{figure*}[t!]
    \centering
    \includegraphics[width=\textwidth]{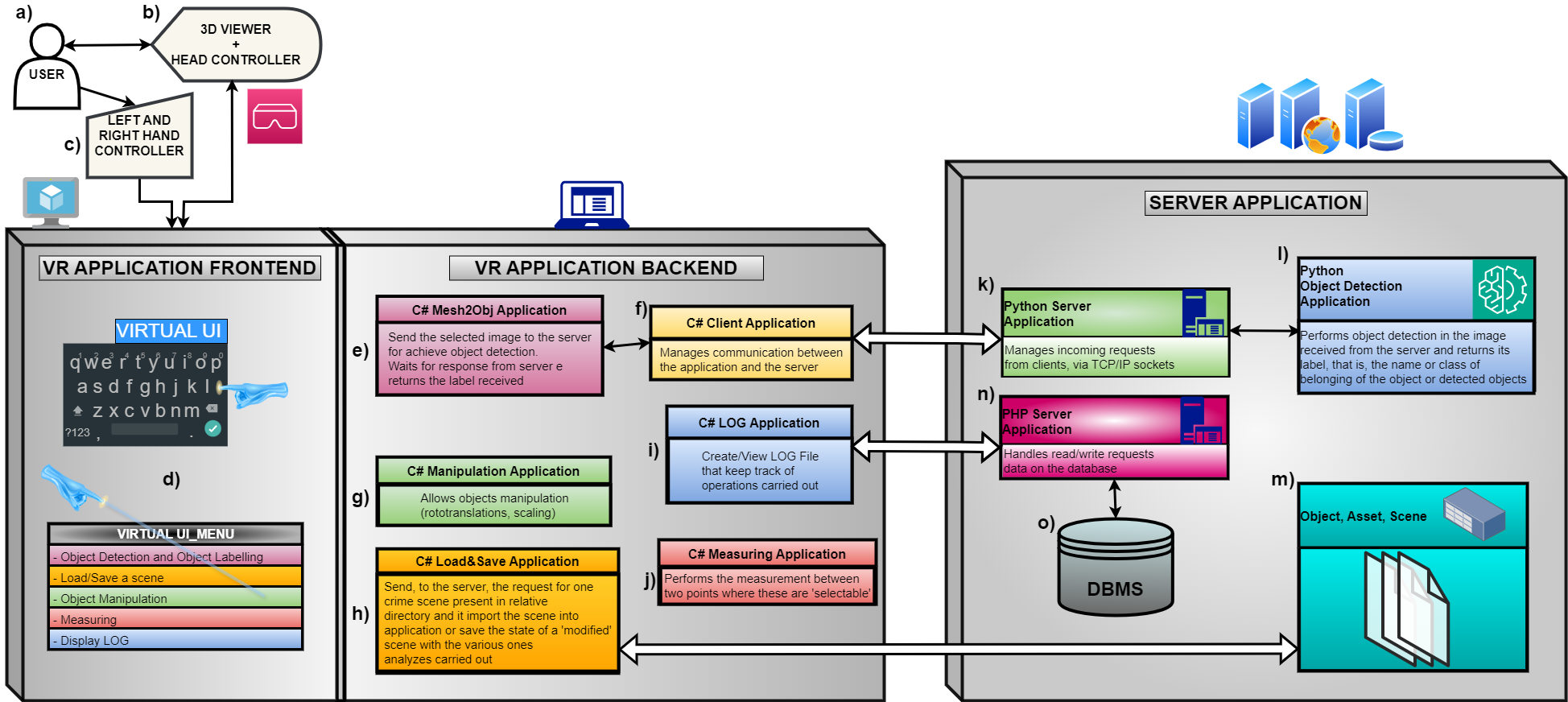}
    \caption{Proposed schema-framework. The virtual reality (VR) application blocks define the immersive environment laboratory (frontend) with related functionalities (employed using the C\# language - backend) to interact with the tools in the scene. The server application includes the Faster-RCNN model for object classification and a database to store the detected objects. The VR application and the Server application communicate via socket.}
    \label{fig:scheme}
\end{figure*}

The analysis of the scene of a crime is a crucial step of the forensic science investigation, which aims to establish criminal responsibility through the collection and examination of traces of evidential value. The scientific literature reported on psychological and emotional influences \cite{casu2023ai} affecting forensic decision-making. Additionally, the need to swiftly identify, isolate and collect traces poses issues related to the preservation and integrity of the scenes. Even at the early stages of a forensic enquiry, it is crucial to minimize ambiguous and unsubstantiated interpretation, as a lack of clarity may compromise the forensic investigation and potentially lead to miscarriages of justice.
Previous research considered the exploitation of 3D techniques to aid forensic scientists and practitioners. Carew et al. \cite{Carew2021} suggested that digitalisation technologies may increase the accuracy and robustness of crime reconstruction in many use cases. Novel 3D imaging methods \cite{Buck2013} and extended reality (XR) technologies hold the potential to record and revisit the scene of a crime minimising contamination \cite{Sieberth2019}, augmenting a wide range of use cases, ranging from training \cite{Wilkins2024} to utilisation in the courtroom \cite{SIEBERTH2021}. Recent studies explored the exploitation of virtual environments and immersive technologies to allow for scene and trace analysis. Guarnera et al. \cite{guarnera2023assessing} developed a virtual laboratory forensic ballistic analysis. The immersive laboratory leveraged a combination of laser scan reconstructions and presented via virtual Reality (VR). This environment provided forensic experts with a virtual laboratory with tools routinely employed in the forensic analysis of cartridge cases and bullets, embedding tools enhanced by the 3D data.

Additionally, the spatial data recorded at the scene allows performing analysis beyond the visual examination. Giudice et al. \cite{giudice2019siamese} proposed a fully automated method based on Siamese Neural Network to compare cartridges to support ballistic analysis, proving the suitability of 3D data, i.e. point clouds, to support the training of machine learning algorithms.

The presented paper capitalises on the exploratory research introduced in the literature, aiming to explore the capabilities of a virtual laboratory by integrating novel deep-learning relying on 2D and 3D imaging methods, eventually allowing for automated analysis, detection, recognition, and labelling of scene objects. Furthermore, a battery of tools for object manipulation and measurement, data recording and exportation to underpin forensic reports and documentation is provided.

Figure~\ref{fig:scheme} shows the diagram of the proposed approach, composed of a VR application and a server application. The former provides the user with an intuitive interface (frontend) to call the methods available on the VR backend. The latter server application exposes a set of APIs through TCP/IP sockets, running the database server and executing intensive tasks such as deep-learning algorithms for object detection.

The main contributions of the proposed work are:
\begin{itemize}
    \item Introduction of a new application for an immersive forensic laboratory to ensure analysis repeatability and the hypotheses testing in a safe environment.
    \item Ability to select portions of the imported spatial data to examine objects of interest.
    \item Implementation of deep-learning techniques in conjunction with immersive virtual reality to enable automated object detection and labelling.
    \item Ability to perform measurements of objects within the virtual crime scene and export examination status.
\end{itemize}

The remainder of this paper is organised as follows. Section~\ref{sec:rworks} outlines state-of-the-art deep learning algorithms for object detection and localization and recent applications of VR Forensics. Section~\ref{sec:dataset} describes the dataset of a simulated crime scene. Then, the proposed framework is presented in the Section~\ref{sec:framework}. Section~\ref{sec:results} provides preliminary detection and localization results, and conclusions are presented in Section~\ref{sec:conclusions}.

\section{Related works}
\label{sec:rworks}

\subsection{Object Detection}
Object detection and localization are tasks that have been widely addressed in the scientific literature of Computer Vision, with particular emphasis on the identification and positioning of objects within images and videos.

In 2014, Girshick et al.~\cite{girshick2014rich} devised the Regions with Convolutional Neural Networks (R-CNN), one of the first frameworks to apply deep learning to object detection. In its original implementation, an R-CNN performed a three-step operation: \textit{(a) Region Proposal}: applying selective search to generate around 2000 candidate regions; \textit{(b) Feature Extraction}: warping each region to a fixed size and processes it with a CNN to extract features, and \textit{(c) Classification}: classifying the extracted features using Support Vector Machines (SVMs).
This framework achieved significant performance improvements over traditional methods, later optimised in \textit{Fast R-CNN}~\cite{girshick2015fast}, which introduced several key innovations: \textit{(a) Single Stage Processing}: processes the image with a CNN to create a feature map; \textit{(b) Region of Interest (RoI) Pooling}: Extracts fixed-length feature vectors from the feature map, and \textit{(c) End-to-End Training}: Trains the CNN, classifier, and bounding box regressor simultaneously with a multitask loss. Further enhancements were introduced by Ren et al.~\cite{ren2016faster} with the framework \textit{Faster R-CNN}, which introduced the concept of Region Proposal Network (RPN) via efficient region proposals, thanks to boundaries prediction and objects score at each location using a set of anchors with different scales and aspect ratios.

In 2016, Liu et al.~\cite{liu2016ssd} proposed a \textit{Single Shot MultiBox Detector} (SSD) that exploits multiscale feature maps to predict bounding boxes and class scores, facilitating the detection of objects of different sizes.

Redmon et al.~\cite{redmon2016you} introduced \textit{You Only Look Once} (YOLO), which treats object detection as a single regression problem. This approach divided the image into a grid predicting bounding boxes and class probabilities directly, enabling real-time detection in a unified architecture. Subsequent versions~\cite{kaur2023comprehensive} further improved accuracy while maintaining real-time performance.
Recently, YOLOv9 \cite{wang2024yolov9} marked a significant advancement in real-time object detection, introducing novel techniques such as Programmable Gradient Information (PGI) and Generalized Efficient Layer Aggregation Network (GELAN). This model demonstrated notable improvements in efficiency, accuracy, and adaptability, establishing new benchmarks on the MS COCO dataset~\cite{lin2014microsoft}.

Lin et al.~\cite{lin2017focal} introduced \textit{RetinaNet}, which improved object recognition in pictures with extreme foreground-background class imbalance by introducing the focal loss function, addressing critical training stages. RetinaNet achieves YOLO-comparable speed whilst also improving detection rate, showing that detectors can be both fast and accurate.

\subsection{Virtual Reality in Forensic Science}
Virtual reality (VR) is a digital technology composing the virtuality continuum defined by Milgram et al. in 1994 \cite{Milgram1994}. More recently, VR has rapidly grown as part of a wider spectrum of technologies, commercially known as extended reality (XR), which encompasses technologies such as Augmented Reality (AR) and Mixed Reality (MR). Immersive virtual reality refers to technologies delivering an enhanced degree of immersion and presence \cite{Slater1997}, commonly delivered through head-mounted displays (HMDs) or CAVE systems. 
Recent studies explored the applicability of this technology in the field of forensic science. 
Despite the latest scanning equipments being  found to deliver the precision accuracy demanded in this field, Maneli et al.\cite{Maneli2022} pointed out that immersive technologies currently suffer from a general lack of trust and are not significantly adopted. 
A combination of high-quality scanning equipment at a reduced cost, in conjunction with the use of real-time rendering engines, demonstrated the feasibility of cost-effective solutions to enable remote practitioners to work together ~\cite{rinaldi2022virtual}. The immersive reconstructions enhanced spatial perception and telepresence, and the ground-truth data available by the software, such as original photographs and reference information, substantiated the collaborative value of the approach ~\cite{rinaldi2022virtual}.
Guarnera et al.~\cite{guarnera2023assessing} devised a VR-based forensic laboratory for cartridge cases and bullets analysis. The use of immersive VR, in combination with unrestricted manipulation of 3D point clouds representing forensic traces, allowed the participants to obtain increased accuracy when performing the ballistics comparison, overcoming the limitations of traditional optical comparator microscopes.

\section{Dataset}
\label{sec:dataset}
\begin{figure}[t!]
    \centering
    \includegraphics[width=.45\textwidth]{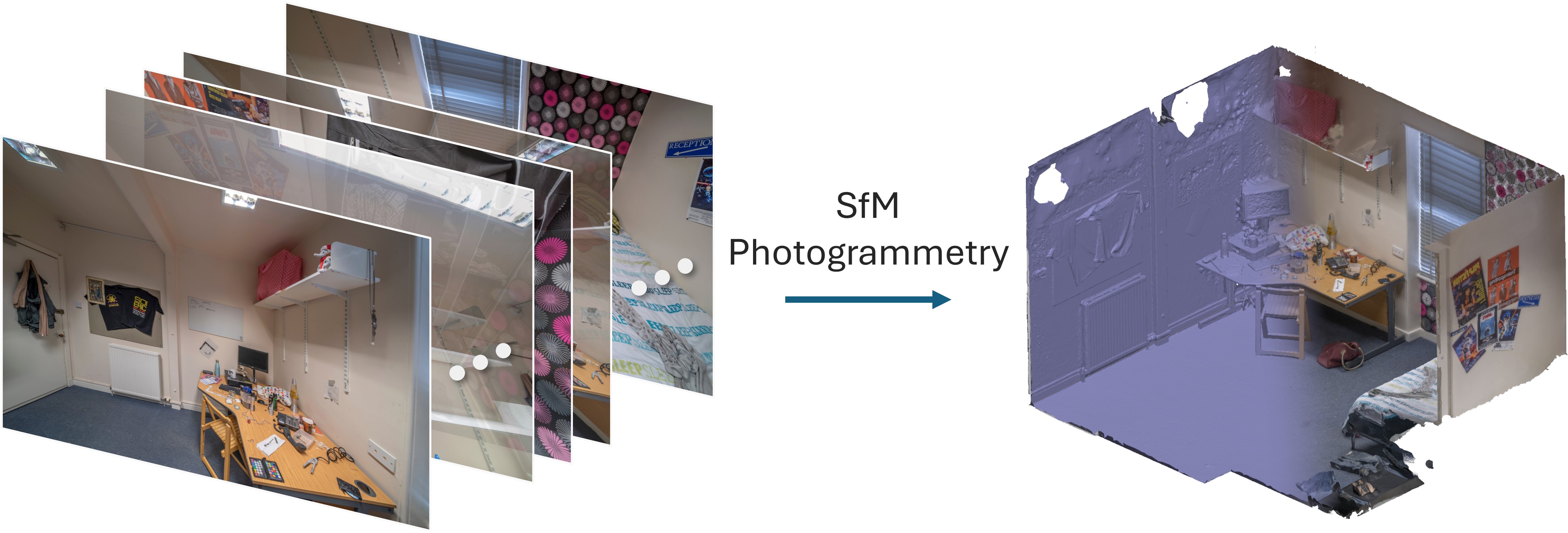}
    \caption{Structure from Motion photogrammetry was used to produce a 3D reconstruction of a simulated scene.}
    \label{fig:dataset}
\end{figure}

The system was tested on a 3D model of a simulated crime scene set in an indoor domestic dwelling. The environment was recorded using the DSLM mirrorless camera Sony ILCE-7SM2, coupled with a Sony FE 12-24 mm F2.8 rectilinear lens mounted on a gimbal panoramic head on a tripod.\\
This setup was used to apply a systematic operating procedure \cite{Yu2023} aimed to optimise the on-site data capture and enable structure-from-motion (SFM) photogrammetry. The resulting dataset consisted of 1044 photographs in RAW format recorded at a focal length of 12 mm, using the aperture priority program employing ambient light only.\\
The images were processed in Adobe Photoshop Camera Raw (version 16.1, Adobe Inc., San Jose, California, USA) and converted into JPG format. The photogrammetric model was computed using Agisoft Metashape software (version 2.0.4.17434, Agisoft LLC, St. Petersburg, Russia) using “Ultra high” quality settings. The SFM software (Figure \ref{fig:dataset}) correctly aligned all the photos and the resulting model was then decimated to a million face count to enable usage in standalone VR headsets. Ultimately, the model was exported as an FBX file to allow importation into the Unity engine.

\section{Proposed Framework}
\label{sec:framework}
The proposed solution was developed using the 3D engine Unity (version 2022.3.0f1 LTS) running on the VR headset Meta Quest 2.
The application consists of 
a VR application, comprising a frontend application that serves as a user interface and a backend component (C\#), which provides a connection to the server application running the deep learning model for object recognition (Python). 
\\
Figure \ref{fig:scheme} shows the main components of the proposed framework:
\begin{enumerate}[a)]
    \item The user interacts with the system via the immersive HMD.
    \item A dedicated script controls the position and orientation of the user in the virtual space.
    \item The user engages with the environment by means of the controllers part of the commercial HMD. Object manipulation and interactions in the user interface are regulated by a dedicated script.
    \item A Virtual User Interface is made available to the user to call the methods exposed by the VR Application Backend.
    \item Method to dispatch an image generated within the VR application to the server to perform object classification.
    \item Method providing communication between the VR application (backend) and the server application.
    \item Method that handles object manipulation, such as translations and rotations.
    \item Method to store and retrieve virtual environment configurations.
    \item Method to write and read operation logs performed by the user within the virtual space.
    \item Measurements method returning the distance between two points.
    \item Endpoint of the TCP/IP socket running on the server to handle requests from the VR application.
    \item Object detection performed by deep learning algorithm of choice.
    \item Location of data (i.e. crime scenes, 3D model assets etc) stored on the server.
    \item PHP Database manager, which handles read and write operations on the database.
    \item Database Management System (DBMS) storing request logs recorded by the server.
\end{enumerate}

\subsection{Framework Features}
\label{sec:subsec}
The proposed framework comprises four main functionalities:

\begin{enumerate}
    \item \textit{Scene Object Detection and Items Labelling:} The initialisation phase prompts the user with the scene selection. Then, a dedicated script manages the detection of scene items and the labelling thereof. This is performed through the vertex selection mode, which allows for selecting a set of mesh vertices of the item to analyse. The collection of vertices is initially vetted to determine if it refers to a closed mesh. Next, an observation plane is created by establishing the position and rotation in the coordinate system of the user's active camera. A refinement phase follows, allowing to expand or shrink the selected vertices. Then, a virtual screenshot is captured, and the generated image is submitted through the TCP/IP socket to perform object classification. The computation is executed on a remote server to cope with the limited computational performance of the standalone headset. 
The image classification algorithm returns a positive result if it successfully identifies any items within the submitted image, returning the class name for the identified items; alternatively, it returns no results and presents a warning to the user. 

\item \textit{Object manipulation:} Object manipulation is achieved by amending the position and orientation of the objects within the virtual scene. This allows for moving and repositioning objects to visually inspect the item of interest from additional points of view and facilitate contextualisation at the site.
Object handling was achieved using the \textit{XR Interaction Toolkit} package, available through the Unity asset manager. This library allows for object interaction and scene navigation through via ray casting, complemented by visual cues to provide intuitive interactions. When manipulating a scene item, a visual reference to the object's original location is maintained to allow accurate repositioning, provided by a duplicate semi-transparent clone of the object at the original location.

\begin{figure}[t!]
    \centering
    \includegraphics[width=.45\textwidth]{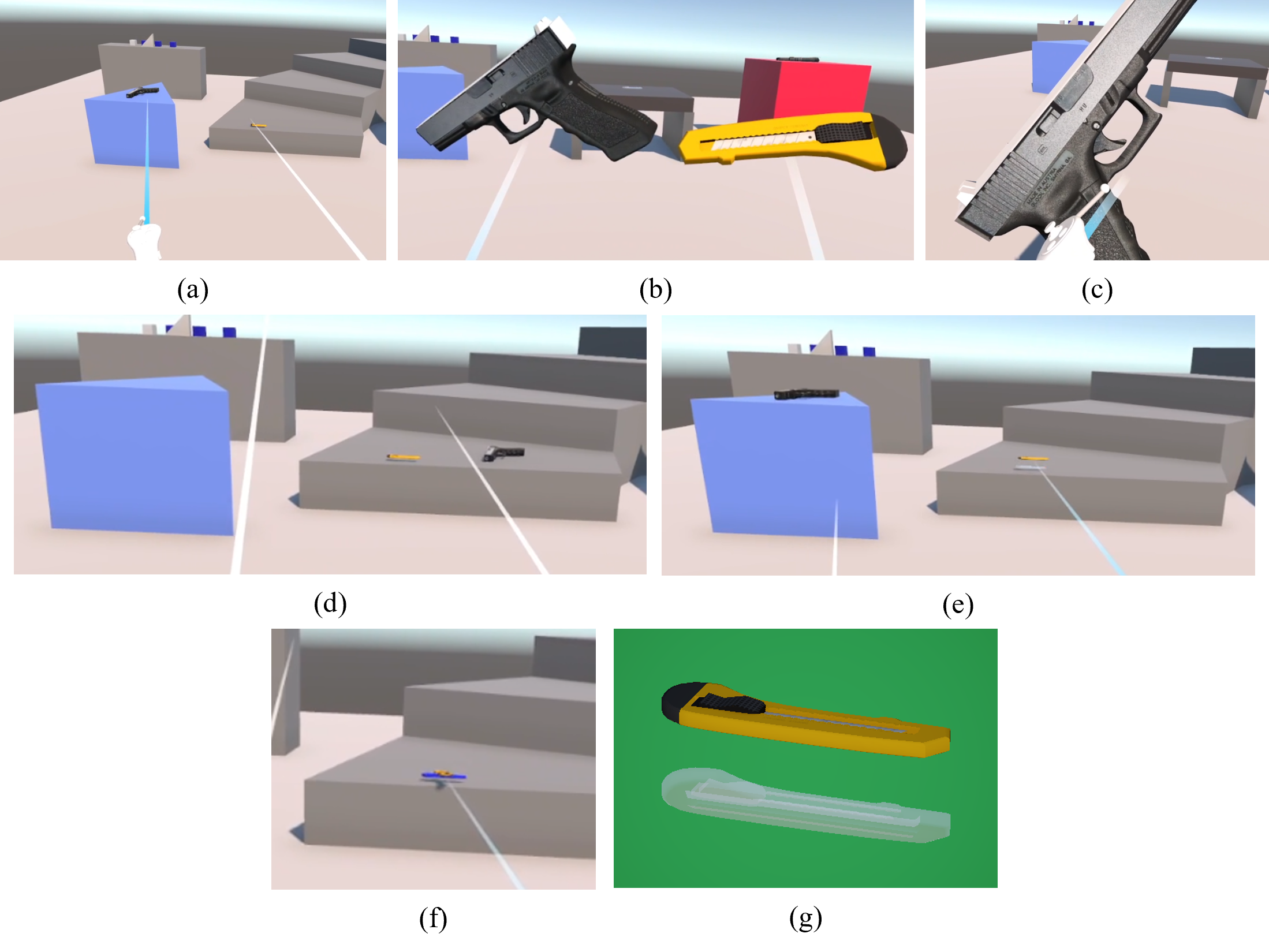}
    \caption{Examples of interactions with objects (N1 = firearm, N2 = cutter). (a) Interaction with object N1, the controller casts a white ray that turns blue when colliding with an interactable object; (b) Visual inspection of objects N1 and N2 performed by rotating and relocating the objects closer to the operator's view; (c) Close inspection of object N1; (d) Release of object N1 in the scene; (e) When interacting with an object, a semi-transparent duplicate indicates its original position; (f) The N2 object is positioned at its original location; (g) Close-up view of object N2 and the semi-transparent placeholder.}
    \label{fig:operations}
\end{figure}

Figure \ref{fig:operations} shows examples of the interaction with objects (item N1: firearm, item N2: cutter). Upon completion of object handling, the user can release the selected item at a new location (Figure \ref{fig:operations}, d), or return it to its original position (Figure \ref{fig:operations}, f). 

\item \textit{Measurement estimation:} Measurement of features of interest is obtained by selecting two points in the virtual space, which constitute the endpoints of the line segment to measure.
The measurement is assessed by computing the magnitude of the Euclidean distance of the two selected points, each represented by a three-dimensional vector. The measurement unit in the virtual scene is equal to 1 metre. 

\item \textit{Creation and management of LOG data:} To enable repeatability and reproducibility of analysis carried out in the immersive laboratory, a LOG registry keeps track of user interactions and requests.  
A complete list of logged actions is stored on the relational database management system (RDBMS) MariaDB version 10.4.11, connecting the frontend application to a server running PHP version 7.4.1 with MySQLi extension.
\end{enumerate}

\section{Experimental Results}
\label{sec:results}
We examined four deep learning models representing the current state of the art pretrained on the COCO dataset \cite{lin2014microsoft}: SSD\footnote{SSD: \url{https://github.com/weiliu89/caffe/tree/ssd?tab=readme-ov-file}, Last Accessed: 10/01/2024}, YOLOv8\footnote{YOLOv8: \url{https://github.com/ultralytics/ultralytics}, Last Accessed: 10/01/2024}, YOLOv9\footnote{YOLOv9: \url{https://github.com/WongKinYiu/yolov9}, Last Accessed: 10/01/2024}, FasterR-CNN\footnote{FasterR-CNN: \url{https://github.com/opencv/opencv_extra/tree/4.x/testdata/dnn}, Last Accessed: 10/01/2024}. The models were not fine-tuned on the dataset examined in this paper. 
Two tests were performed to evaluate the performance of the selected methods. Firstly, we benchmark each deep learning model against the set of images that generated the SFM reconstruction by measuring the overall number of identified objects and their accuracy score. Secondly, each DL model was tasked with labeling the image provided by the user, as explained in section \ref{sec:subsec}1.
The models were trained using Intel® Xeon® Silver 4214R and Nvidia RTX A6000 (48 GB).

\subsection{Detection and Localisation on SFM Images}
\begin{figure}
    \centering
    \includegraphics[width=.5\textwidth]{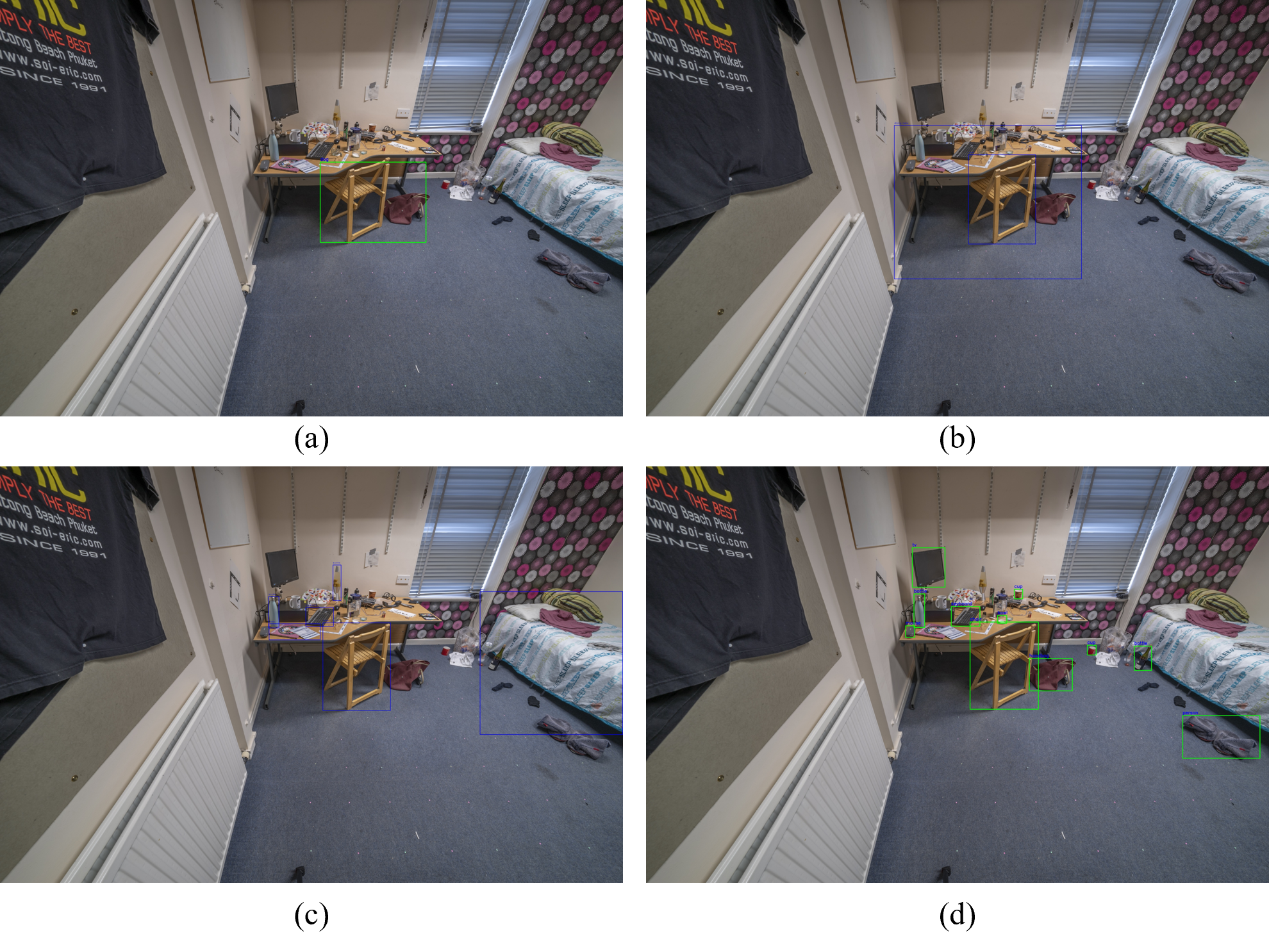}
    \caption{Examples of object classification preformed by (a) SSD; (b) YOLO-v8; (c) YOLO-v9; (d) Faster-RCNN.}
    \label{fig:rsults}
\end{figure}

We tested the models by considering the following metrics: the number of detected objects, classification accuracy, and the average of the computational time. The image recognition was performed on a random subset of 100 images. 
Figure~\ref{fig:rsults} shows an example of the visual localization and detection results. The experimental results are shown in Table~\ref{tab:res1}. 

\begin{table}
\caption{Results of object detection algorithms.}
\begin{tabular}{cccc}
\cline{2-4} & \textbf{\begin{tabular}[c]{@{}c@{}}\#Detected\\ Object\end{tabular}} & \textbf{\begin{tabular}[c]{@{}c@{}}Accuracy (\%)\end{tabular}} & \textbf{\begin{tabular}[c]{@{}c@{}} Avg Computational \\ Time (sec)\end{tabular}}  \\ \hline
\textbf{SSD} & 82 & \textbf{74} & 300 \\ \hline
\textbf{YOLOv8}  & 250  & 68  & \textbf{60} \\ \hline
\textbf{YOLOv9} & 894  & 72 & 420  \\ \hline
\textbf{FasterR-CNN} & \textbf{2098} & 72& 1080\\ \hline
\end{tabular}
\label{tab:res1}
\end{table}

The results show that the FasterR-CNN model is able to detect most objects, with a much greater identification rate. Regarding accuracy, SSD performed marginally better than YOLOv8 and FasterR-CNN. YOLOv8 was notably faster than other models, to the detriment of the detection rate. FasterR-CNN was deemed the most suitable model due to its high rate of identification compared to other models, despite the higher computational time required to analyse the whole set of images.

\subsection{Detection and Localisation of Specific Objects}

\begin{table}[h]
\caption{Detection and localization of specific objects. The classification accuracy value (\%) is reported. X indicates that the specific object was not detected. Red text indicates an incorrect classification (we also report the predicted class).}
\begin{tabular}{cccccc}
\hline
\textbf{} & \textbf{SSD} & \textbf{YOLOv8} & \textbf{YOLOv9} & \textbf{FasterR-CNN} \\ \hline
\textbf{tv} & X & X & X & \textbf{98.90} \\ \hline
\textbf{cup} & X & X & X & \textbf{91} \\ \hline
\textbf{keyboard} & X & X & 73 & \textbf{94} \\ \hline
\textbf{bed} & X & X & \textbf{89} & X \\ \hline
\textbf{bottle} & X & X & 58 & \textbf{79} \\ \hline
\textbf{chair} & {\color[HTML]{FE0000} 83 (dog)} & 56 & \textbf{76} & 72 \\ \hline
\textbf{dining-table} & X & \textbf{69} & X & X \\ \hline
\textbf{person} & X & X & X & {\color[HTML]{FE0000} 67 (dress)} \\ \hline
\textbf{book} & X & X & \textbf{63} & X \\ \hline
\textbf{handbag} & X & X & X & \textbf{59} \\ \hline
\textbf{bowl} & X & X & X & \textbf{53} \\ \hline
\end{tabular}
\label{tab:my-table}
\end{table}
The second evaluation focused on the detection, localisation and labelling of objects within the selected portion of an image provided by the user. Each object was extracted as described in Section~\ref{sec:framework}. Table~\ref{tab:my-table} shows the results obtained in the detection and localization of the following objects: TV, cup, keyboard, bed, bottle, chair, dining table, person, book, handbag, and bowl. 


SSD was unable to correctly identify any of the items of interest, additionally, incorrectly recognised one of the labels. YOLOv9 achieved better results compared to YOLOv8 by correctly returning four labels (keyboard, bed, bottle, book), with higher accuracy for one label (chair), yet being unable to identify one (dining table). FasterR-CNN proved to be the best model overall, being able to identify seven labels with varying degrees of accuracy, although erroneously identifying one object type ("dress" instead of "person"). 

\section{Conclusion, Discussion and Future Works}
\label{sec:conclusions}
The meticulous analysis of crime scenes is crucial in forensic investigations, requiring precision, adherence to best practices, and critical thinking from forensic investigators. The need to produce accurate and unbiased reports that meet judicial standards highlights the complexity of this task. Additionally, the dynamic nature of crime scenes, which are susceptible to deterioration and contamination, presents significant challenges in documenting and isolating evidential traces.
To address this task, we proposed a novel approach using photogrammetric reconstruction and virtual reality technology, along with autonomous object recognition powered by a pre-trained FasterR-CNN model, aimed at improving the outcomes of crime scene analysis. Experimental results show the feasibility and effectiveness of this method in identifying and analysing objects with evidentiary value while reducing subjective bias and contamination risks.

The combination of VR, DL techniques and client-server architecture is complex to manage and requires specific skills to use. However, it emphasises the potential benefits, such as increased accuracy and efficiency in crime scene analysis, which justify the initial investment in skills and resources.

Future work will focus on the analysis of multiple crime scenes, for improving and training new deep-learning models and allowing detection of new objects and traces characteristic of crime scenes, such as blood and fire patterns. Detailed studies on forensic experiments will also be included to assess usability and effectiveness.

\bibliographystyle{IEEEtran.bst}
\balance{
\bibliography{main}

\begin{thebibliography}{10}
\providecommand{\url}[1]{#1}
\csname url@samestyle\endcsname
\providecommand{\newblock}{\relax}
\providecommand{\bibinfo}[2]{#2}
\providecommand{\BIBentrySTDinterwordspacing}{\spaceskip=0pt\relax}
\providecommand{\BIBentryALTinterwordstretchfactor}{4}
\providecommand{\BIBentryALTinterwordspacing}{\spaceskip=\fontdimen2\font plus
\BIBentryALTinterwordstretchfactor\fontdimen3\font minus \fontdimen4\font\relax}
\providecommand{\BIBforeignlanguage}[2]{{%
\expandafter\ifx\csname l@#1\endcsname\relax
\typeout{** WARNING: IEEEtran.bst: No hyphenation pattern has been}%
\typeout{** loaded for the language `#1'. Using the pattern for}%
\typeout{** the default language instead.}%
\else
\language=\csname l@#1\endcsname
\fi
#2}}
\providecommand{\BIBdecl}{\relax}
\BIBdecl

\bibitem{casu2023ai}
M.~Casu, L.~Guarnera, P.~Caponnetto, and S.~Battiato, ``Ai mirage: The impostor bias and the deepfake detection challenge in the era of artificial illusions,'' \emph{arXiv preprint arXiv:2312.16220}, 2023.

\bibitem{Carew2021}
R.~M. Carew, J.~French, and R.~M. Morgan, ``{3D forensic science: A new field integrating 3D imaging and 3D printing in crime reconstruction},'' \emph{Forensic Science International: Synergy}, vol.~3, p. 100205, jan 2021.

\bibitem{Buck2013}
\BIBentryALTinterwordspacing
U.~Buck, S.~Naether, B.~R{\"{a}}ss, C.~Jackowski, and M.~J. Thali, ``{Accident or homicide – Virtual crime scene reconstruction using 3D methods},'' \emph{Forensic Science International}, vol. 225, no. 1-3, pp. 75--84, feb 2013. [Online]. Available: \url{https://linkinghub.elsevier.com/retrieve/pii/S0379073812002587}
\BIBentrySTDinterwordspacing

\bibitem{Sieberth2019}
\BIBentryALTinterwordspacing
T.~Sieberth, A.~Dobay, R.~Affolter, and L.~C. Ebert, ``{Applying virtual reality in forensics – a virtual scene walkthrough},'' \emph{Forensic Science, Medicine and Pathology}, vol.~15, no.~1, pp. 41--47, mar 2019. [Online]. Available: \url{http://link.springer.com/10.1007/s12024-018-0058-8}
\BIBentrySTDinterwordspacing

\bibitem{Wilkins2024}
H.~V. Wilkins, V.~Spikmans, R.~Ebeyan, and B.~Riley, ``{Application of augmented reality for crime scene investigation training and education},'' \emph{Science {\&} Justice}, vol.~64, no.~3, pp. 289--296, may 2024.

\bibitem{SIEBERTH2021}
\BIBentryALTinterwordspacing
T.~Sieberth, D.~Seckiner, A.~Dobay, E.~Dobler, R.~Golomingi, and L.~Ebert, ``The forensic holodeck – recommendations after 8 years of experience for additional equipment to document vr applications,'' \emph{Forensic Science International}, vol. 329, p. 111092, 2021. [Online]. Available: \url{https://www.sciencedirect.com/science/article/pii/S0379073821004126}
\BIBentrySTDinterwordspacing

\bibitem{guarnera2023assessing}
L.~Guarnera, O.~Giudice, S.~Livatino, A.~B. Paratore, A.~Salici, and S.~Battiato, ``Assessing forensic ballistics three-dimensionally through graphical reconstruction and immersive vr observation,'' \emph{Multimedia Tools and Applications}, vol.~82, no.~13, pp. 20\,655--20\,681, 2023.

\bibitem{giudice2019siamese}
O.~Giudice, L.~Guarnera, A.~B. Paratore, G.~M. Farinella, and S.~Battiato, ``Siamese ballistics neural network,'' in \emph{2019 IEEE International Conference on Image Processing (ICIP)}.\hskip 1em plus 0.5em minus 0.4em\relax IEEE, 2019, pp. 4045--4049.

\bibitem{girshick2014rich}
R.~Girshick, J.~Donahue, T.~Darrell, and J.~Malik, ``Rich feature hierarchies for accurate object detection and semantic segmentation,'' in \emph{Proceedings of the IEEE Conference on Computer Vision and Pattern Recognition}, 2014, pp. 580--587.

\bibitem{girshick2015fast}
R.~Girshick, ``Fast r-cnn,'' in \emph{Proceedings of the IEEE International Conference on Computer Vision}, 2015, pp. 1440--1448.

\bibitem{ren2016faster}
S.~Ren, K.~He, R.~Girshick, and J.~Sun, ``Faster r-cnn: Towards real-time object detection with region proposal networks,'' \emph{IEEE Transactions on Pattern Analysis and Machine Intelligence}, vol.~39, no.~6, pp. 1137--1149, 2016.

\bibitem{liu2016ssd}
W.~Liu, D.~Anguelov, D.~Erhan, C.~Szegedy, S.~Reed, C.-Y. Fu, and A.~C. Berg, ``Ssd: Single shot multibox detector,'' in \emph{Computer Vision--ECCV 2016: 14th European Conference, Amsterdam, The Netherlands, October 11--14, 2016, Proceedings, Part I 14}.\hskip 1em plus 0.5em minus 0.4em\relax Springer, 2016, pp. 21--37.

\bibitem{redmon2016you}
J.~Redmon, S.~Divvala, R.~Girshick, and A.~Farhadi, ``You only look once: Unified, real-time object detection,'' in \emph{Proceedings of the IEEE conference on computer vision and pattern recognition}, 2016, pp. 779--788.

\bibitem{kaur2023comprehensive}
R.~Kaur and S.~Singh, ``A comprehensive review of object detection with deep learning,'' \emph{Digital Signal Processing}, vol. 132, p. 103812, 2023.

\bibitem{wang2024yolov9}
C.-Y. Wang and H.-Y.~M. Liao, ``{YOLOv9}: Learning what you want to learn using programmable gradient information,'' 2024.

\bibitem{lin2014microsoft}
T.-Y. Lin, M.~Maire, S.~Belongie, J.~Hays, P.~Perona, D.~Ramanan, P.~Doll{\'a}r, and C.~L. Zitnick, ``Microsoft coco: Common objects in context,'' in \emph{Computer Vision--ECCV 2014: 13th European Conference, Zurich, Switzerland, September 6-12, 2014, Proceedings, Part V 13}.\hskip 1em plus 0.5em minus 0.4em\relax Springer, 2014, pp. 740--755.

\bibitem{lin2017focal}
T.-Y. Lin, P.~Goyal, R.~Girshick, K.~He, and P.~Doll{\'a}r, ``Focal loss for dense object detection,'' in \emph{Proceedings of the IEEE International Conference on Computer Vision}, 2017, pp. 2980--2988.

\bibitem{Milgram1994}
P.~Milgram and F.~Kishino, ``{A Taxonomy of Mixed Reality Visual Displays},'' \emph{IEICE Transactions on Information Systems}, vol. E77-D, no.~12, pp. 1--15, 1994.

\bibitem{Slater1997}
\BIBentryALTinterwordspacing
M.~Slater and S.~Wilbur, ``{A framework for immersive virtual environments (FIVE): Speculations on the role of presence in virtual environments},'' \emph{Presence: Teleoperators and Virtual Environments}, vol.~6, no.~6, pp. 603--616, dec 1997. [Online]. Available: \url{https://direct.mit.edu/pvar/article/6/6/603-616/18157}
\BIBentrySTDinterwordspacing

\bibitem{Maneli2022}
M.~A. Maneli and O.~E. Isafiade, ``{3D Forensic Crime Scene Reconstruction involving Immersive Technology: A Systematic Literature Review},'' \emph{IEEE Access}, vol.~10, pp. 88\,821--88\,857, 2022.

\bibitem{rinaldi2022virtual}
V.~Rinaldi, L.~Hackman, and N.~NicDaeid, ``Virtual reality as a collaborative tool for digitalised crime scene examination,'' in \emph{International Conference on Extended Reality}.\hskip 1em plus 0.5em minus 0.4em\relax Springer, 2022, pp. 154--161.

\bibitem{Yu2023}
\BIBentryALTinterwordspacing
S.-h. Yu, G.~Thomson, V.~Rinaldi, C.~Rowland, and N.~N. Daeid, ``{Development of a Dundee Ground Truth imaging protocol for recording indoor crime scenes to facilitate virtual reality reconstruction},'' \emph{Science {\&} Justice}, vol.~63, no.~2, pp. 238--250, mar 2023. [Online]. Available: \url{10.1016/j.scijus.2023.01.001}
\BIBentrySTDinterwordspacing

\end{thebibliography}
}
\end{document}